\documentclass[fullpaper,cameraready]{nldl}
\usepackage[square,numbers]{natbib}
\renewcommand{\DOI}{10.7557/18.6257}
\usepackage[utf8]{inputenc}
\usepackage{url}
\usepackage{graphicx}
\usepackage{authblk}
\usepackage{listings}
\usepackage{comment}
\usepackage{amsmath}
\usepackage{amsfonts}
\usepackage{listings}
\usepackage[frozencache,cachedir=.]{minted}

\usepackage{tikz,graphics,color,fullpage,float,epsf,caption,subcaption}

\usepackage{algorithm}
\usepackage{algorithmic}

\usepackage{hyperref}

\newcommand{\tangh}{\textsc{tanh}}
\newcommand{\sigmoid}{\textsc{sigmoid}}
\newcommand{\relu}{\textsc{relu}}
\newcommand\figref[1]{Fig.~\ref{#1}}
\newcommand\secref[1]{Sec.~\ref{#1}}

\title{Mutual information estimation for graph convolutional neural networks}

\author[1]{Marius C. Landverk}
\author[2]{Signe Riemer-S\o{}rensen}
\affil[1]{mariuslandc@gmail.com}
\affil[2]{signe.riemer-sorensen@sintef.no, Analytics and AI, SINTEF Digital, P.O. Box 124 Blindern, N-0314 Oslo, Norway}

\date{\vspace{-5ex}}

\begin{document}
\nldlmaketitle

\begin{abstract}
Measuring model performance is a key issue for deep learning practitioners. However, we often lack the ability to explain why a specific architecture attains superior predictive accuracy for a given data set. Often, validation accuracy is used as a performance heuristic quantifying how well a network generalizes to unseen data, but it does not capture anything about the information flow in the model. 

Mutual information can be used as a measure of the quality of internal representations in deep learning models, and the information plane may provide insights into whether the model exploits the available information in the data. 

The information plane has previously been explored for fully connected neural networks and convolutional architectures. We present an architecture-agnostic method for tracking a network's internal representations during training, which are then used to create the mutual information plane. The method is exemplified for graph-based neural networks fitted on citation data. We compare how the inductive bias introduced in graph-based architectures changes the mutual information plane relative to a fully connected neural network. 
\end{abstract}

\section{Introduction}
\subsection{Motivation}
For classification problems, the validation accuracy is a common heuristic to gauge the generalization capabilities of a model. Whilst it is a useful metric, it leaves something to be desired in terms of understanding why a model is able to perform as well, or as ill, as it does. 
The ability of a model to fit the data is directly related to the quality of the representations generated by the model. The information plane is given as the mutual information between a model's representations, $Z_i$, with the input $X$ and the true labels $Y$~\cite{info_bottleneck_original, tishby2015deep, openingblackbox}. Visual inspection of the information plane at the point in training where the performance plateaus can provide qualitative guidance. In this paper, we seek to use the information plane from the information bottleneck method~\cite{openingblackbox} in order to gauge model fit across architectures with different inductive biases, and we exemplify by comparing a fully-connected neural network to graph-based networks trained on the same classification task.

\subsection{Mutual information}
The core of the information bottleneck method is a quantification of shared information between random variables originally suggested by~\cite{info_bottleneck_original}. For two random variables $X$ and $Y$, the mutual information $I(X, Y)$ is a symmetric quantity measuring the amount of information that $X$ contains about $Y$, and vice versa. If we want to predict the true labels $Y$ from $X$, we can define a compressed version of $X$, called $Z$, with shared information $I(X, Z)$ and $I(Z, Y)$. The minimal representation $Z$ (maximally compressed information on $Y$ from $X$), that simultaneously maximises the mutual information with $Y$ is called the 'information bottleneck'~\cite{openingblackbox, NonlinearIB}:
\begin{equation}
    \label{eqn:IBVariable-def}
    \mathrm{argmax}_{Z \in \Delta} I(Y, Z) \; \mathrm{such\, that} \; I(X, Z) \leq R,
\end{equation}
where $R$ is a given scalar threshold. Mutual information can be expressed in terms of the entropy $H(X), \; H(Y)$, as $I(X, Y) = H(X) - H(X|Y) = H(Y) - H(Y|X)$, where $H(Y|X)$ is the conditional entropy of $Y$ given $X$~\cite{elements_of_information_theory}. 

If $Y, X$ and $Z$ are random variables in a Markov chain where $Y \rightarrow X \rightarrow Z$, the data-processing inequality applies to the mutual information between the variables ~\cite{elements_of_information_theory}:
\begin{equation}
    \label{eqn:dpi-inequality}
    I(X, Y) \geq I(Z, Y).
\end{equation}

In other words, the information about an input $X$ contained in $Z$ cannot be increased by applying a function to $Z$, as long as the function does not use additional information about $X$. 

For neural networks, the representations $Z_i$ can be associated with the successive hidden layers. For fully connected neural networks, the data processing inequality can then be applied to obtain a set of inequalities. However,  in more complicated neural network structures such as recurrent neural networks or graph convolutional neural networks, the hidden layers $Z$ draws on information about $X$, and the data processing inequality is broken.
In the case of graph convolutional neural network, the data processing inequality is broken explicitly because all hidden layers uses the edge information in $X$ to compute the forward pass.

\subsection{Related work}
It has been debated whether the mutual information follows a specific pattern during training with a fitting phase followed by a compression phase~\cite{openingblackbox, harvardinfothry}. So far, the compression phase is only observed for symmetric activation functions such as \sigmoid{} and \tangh{} (see~\cite{2020arXiv200309671G} for an overview of experiments). Here we do not make any claims regarding the compression phase, but provide a framework for comparing the mutual information across neural network architectures.

Early works~\cite{tishby2015deep, openingblackbox} use a binning procedure to estimate the mutual information. This is only meaningful for activation functions with restricted output range e.g. \sigmoid, and the choice of binning has been proven to affect the resulting mutual information estimate significantly~\cite{2021arXiv210612912S}. This can be alleviated by assuming a distribution for the hidden variables $Z$ and rewriting the mutual information to only include the hidden layer representations. In \secref{sec:method} we focus on this method which allows to derive both upper and lower bounds for $I(X, Z)$ and $I(Z, Y)$~\cite{pairwise_distance_kolchinsky, NonlinearIB, harvardinfothry}, and only briefly comment here on alternative methods. Instead of assuming the distributions, the approach of~\cite{mine_paper} relies on a completely separate neural network to estimate the mutual information of the problem-specific neural network. However, this introduces additional hyper parameters and high risk of numerical instabilities.

The method of~\cite{RENYIMIPAPER} uses an estimate for the entropy based on an eigenvalue decomposition of $X$ and a smoothing with an infinitely-divisible, real-valued positive definite kernel. However, since the method relies on computing eigenvalues, the computational requirements fast become prohibitive, as this computation is done several times in order to obtain a single mutual information plot.

The authors of~\cite{dissecting-deep-networks-mi} apply the mutual information plane as a way to explain convolutional neural networks trained for image classification (e.g. ResNET), but they only consider that specific architecture and image-like data.

\subsection{Our contribution}
We present a framework for tracking the activations of a generic neural network. To produce an information plane, a classification task is required, since otherwise the conditional entropy $H(X|Y)$ becomes intractable. The framework has been developed in PyTorch~\cite{pytorch} and can be found on GitHub\footnote{\url{https://github.com/mariusmcl/info-bottleneck-tracking}}.

The novelty of the developed framework is that it is agnostic to the kind of model architecture, as long as all the submodules to be tracked are defined explicitly in the initialization method of the PyTorch module. % Instead of a nn.Linear layer, one could have a nn.Conv layer, or a nn.RNN layer. 
The original methods for mutual information estimations were developed on fully connected neural networks. To our knowledge, mutual information has not previously been estimated for graph neural networks or recurrent networks (see e.g. table 1 in~\cite{2020arXiv200309671G}).
The already existing codebases from~\cite{info_bottleneck_original, harvardinfothry} are well suited for the specifically investigated network architectures and setup, but they lack the flexibility to be easily applied on different neural architectures, and hence do not invite comparisons across architectures.

In \secref{sec:examples}, we exemplify on graph-like data and discuss how the inductive bias introduced by the imposed structure of the graph neural network affects the model quality and how that is expressed in the mutual information plane.

\section{Method} \label{sec:method}
We estimate the mutual information based on the method from Kolchinsky et. al.~\cite{pairwise_distance_kolchinsky, NonlinearIB}. Given a batch $B$ with $N_B$ samples, the upper bound estimates\footnote{The lower bound is obtained similarly by replacing the 2 in $K_{ij}$ with 8.} for the mutual information $I(X, Z_B)$ and $I(Z_B, Y)$ are 

\begin{equation}
    \label{eqn:kolchinsky-IXT-lowebound}
    I(X, Z_B) \leq -\frac{1}{N_B}\sum_i \log \frac{1}{N_B}\sum_j K_{ij}
\end{equation}
with $K_{ij} = \exp \left( -\|Z_i - Z_j\|_2^2/(2\sigma^2) \right)$, and

\begin{align}
    \label{eqn:kolchinsky-ITY-lowebound}
    I(Z_B, Y)&\leq  -\frac{1}{N_B}\sum_i \log \frac{1}{N_B}\sum_j K_{ij} \\ \nonumber
    &-\sum_{c=1}^C p_c \left[-\frac{1}{N_c}\sum_{\{i | Y_i=c\}} \log \frac{1}{N_c} \sum_{\{j | Y_j=c\}} K_{ij} \right],
\end{align}
with $p_c=\frac{N_c}{N}$ where $N_c$ is the total number of examples in category $c$ and $N$ is the total number of samples.

The estimate is based on pairwise distances between $Z_i$ and $Z_j$. If the components in the representation of a class (some/all classes) are very clustered, their pairwise distances are small and hence the second term in equation~\eqref{eqn:kolchinsky-ITY-lowebound} becomes small, leading to large mutual information. If the representation is e.g. random and thereby not clustered at all, the mutual information becomes large. Hence, if the inter-class distances are small, one can relate a sub-region of the representation's domain to a class label, which in turn would yield a higher estimate of mutual information.

The only hyperparameter in the method is the homoscedastic noise variance $\sigma^2$ of the assumed normal distribution. In equations~\eqref{eqn:kolchinsky-IXT-lowebound} and~\eqref{eqn:kolchinsky-ITY-lowebound}, $\sigma^2$ can be interpreted as a bandwidth term, with a higher $\sigma^2$ allowing more interactions between batch elements $i$ and $j$, leading to higher values of $K_{ij}$. Different choices of $\sigma^2$ produce quantitatively and qualitatively different results. If the variance is too small, the estimates become noisy (large scatter in the plots), and if it becomes too big the estimates have large uncertainty, which also tends to break the data processing inequality even for fully connected neural networks. This is consistent with previous observations based on independent estimators~\cite{2020arXiv200309671G, 2018arXiv180400057Y, 2018arXiv180406537Y}.  

The developed framework is designed to be compatible with the PyTorch training workflow and relies on recording the activation values of the model during training by attaching a "hook" on the forward pass\footnote{Pytorch: Forward And Backward Function Hooks, \url{https://pytorch.org/tutorials/beginner/former_torchies/nnft_tutorial.html##forward-and-backward-function-hooks}}. We show an example training loop with activation tracking during training in code listing \ref{code:framework-training-loop}. The forward hooks are defined through the model's \texttt{named}\_\texttt{children} attribute\footnote{Pytorch: Modules \url{https://pytorch.org/docs/stable/generated/torch.nn.Module.html}}.

\begin{listing}[ht]
\begin{minted}{python}
for epoch in range(num_epochs):
    hooks = list(tracker.forward_hooks.keys())
    tracker.register_new_epoch(hooks)
    for x, y in train_loader:
        optimizer.zero_grad()
        out = classifier(x)
        loss = crossentropy_loss(out, y)
        loss.backward()
        optimizer.step()
        tracker.save()
\end{minted}
\caption{An example of activation tracking during a training loop in PyTorch for training a network for a classification task. The forward hooks are defined through the model's $\text{named}\_\text{children}$ attribute.}
\label{code:framework-training-loop}
\end{listing}

We have verified that the method reproduces the mutual information estimates of~\cite{openingblackbox} and~\cite{dissecting-deep-networks-mi} (as far as possible with the details provided in the paper).

\section{Example on graph-like data} \label{sec:examples}
We exemplify the method on two different graph-like datasets (\secref{sec:cora}, \secref{sec:arxiv}) and and three different architectures (\secref{sec:mlp}-\secref{sec:gat}).

\subsection{The Cora data} \label{sec:cora}
The Cora data\footnote{\url{https://relational.fit.cvut.cz/dataset/CORA}} has a graph like structure with 2708 nodes representing scientific papers and 5429 edges given by citations between papers. The edges are unidirectional so paper A can cite paper B or opposite, or they can both cite each other. The papers are described by a feature vector of length $D=1433$ where each element is the number count of a predetermined word (bag-of-words feature vector). In addition, each of the papers belong to one of seven categories indicating which scientific field the article is published in.
For the Cora data, we consider the task of classifying the papers into one of seven publication categories.
We use the built in masks for splitting in training (140 nodes), validation (500 nodes) and test (1000 nodes) samples.

\subsection{The arxiv data} \label{sec:arxiv}
The arxiv dataset\footnote{\url{https://ogb.stanford.edu/docs/nodeprop/\#ogbn-arxiv}} contains the citation network between preprint papers in computer science submitted to arXiv. The nodes represents 169,343 preprints connected by 1,166,243 directed edges representing one preprint citing another. The papers are described by a feature vector of length $D=128$ where each element is the number count of a predetermined word (bag-of-words feature vector). The targets are one of 40 sub-categories assigned to each paper (e.g. cs.AI, cs.LG etc. which are manually assigned by the authors and moderators), and hence the task is classification. We follow the recommendation and split into training and test nodes based on submission dates such that we train on papers published until 2017, validate on those published in 2018, and test on those published since 2019. This is known to be slightly more challenging than the random split used in the Cora data~\cite{2020arXiv200500687H}.

\subsection{Multilayer perceptron model} \label{sec:mlp}
The simplest model is a vanilla multilayer perceptron (MLP) model~\cite{deep-learning-book-main}. The propagation rule for the MLP is given by
\begin{equation}
    \label{eqn:fcn-update-rule}
    Z_{l+1} = \sigma\bigl( Z_l W_{l+1} \bigr),
\end{equation}

where $W$ is the weight matrix, and $\sigma$ indicates the activation function. The MLP was chosen as reference to explore if the mutual information plane is affected by inductive bias in the model.

\subsection{Graph convolution model} \label{sec:gcn}
Graph convolutions are a generalisation of the convolutionary operator that have become a standard workhorse in deep learning on images, to operate on arbitrary graphs~\cite{GCNConv, 2018arXiv180601261B}. The propagation rule for a graph convolutional network is achieved through a weighted sum of neighboring nodes' features, followed by a matrix multiplication. The propagation rule for a node $n$ is given by
\begin{equation}
    \label{eqn:graph-update-rule}
    Z_{l+1}^n = \sigma \left( \left(\sum_{j \in \mathcal{N}(n)} \frac{1}{\sqrt{\hat{d}_j\cdot \hat{d}_n}}Z_{l}^j\right)W_{l+1}  \right),
\end{equation}
with $\hat{d}_i$ being defined as $\hat{d}_i = 1 + \sum_{j \in \mathcal{N}(i)} e_{ij}$ with $e_{ij}$ being the edge weight between nodes $i$ and $j$, with the default being $e_{ij}=1$. The graph convolutional architecture thus has a fixed weighing of a nodes' attribute based on that nodes' edge degree. 

\subsection{Graph attention model} \label{sec:gat}
Graph attention networks follows the same propagation rule as in equation~\eqref{eqn:graph-update-rule} but apply an attention mechanism at the point when the information is aggregated from the neighbouring nodes through normalised attention scores assigned to each source node before the summation~\cite{2017arXiv171010903V}. For a single node $n$ the propagation can be written as:
\begin{equation}
    \label{eqn:gat-update-rule}
    Z_{l+1}^n = \sigma \left( \left(\sum_{j \in \mathcal{N}(n)} \alpha^{j, n}Z_{l}^j\right)W_{l+1}  \right) \, ,
\end{equation}
where the attention scores $\alpha^{j, n}$ are trainable and thus not fixed as for the graph convolutional network.

\subsection{Training the models}
All architectures are fitted using three hidden layers, with 300, 200 and 100 neurons each. For the Cora data we use \relu{} activations, while for the arxiv data we fit two models for each architecture using \tangh{} and \relu{} activations respectively. We train for 600 epochs for the arxiv dataset, and for 100 epochs for the Cora dataset. Due to the difficulties of batching a graph dataset, all models are trained with full-batch gradient descent, and as such each epoch corresponds to one update of the model parameters.

We estimate the upper bound on the mutual information using Listing \ref{code:framework-training-loop} with a noise parameter of $\sigma^2 = 0.1$ for the Cora dataset and $\sigma^2 = 0.001$ for the arxiv dataset. We have tested multiple values of $\sigma$. Generally, we find that small values ($\sigma=10^{-5}$) lead to very scattered values of mutual information, while large values ($\sigma=0.1$) reduce the scatter at the cost of increased uncertainty, which may lead to apparent violation of the data processing inequality for the MLP model on some datasets. We selected the values of $\sigma$ such that the data processing inequality was not (strongly) violated for the MLP model (see also \secref{sec:method}).

\subsection{Results on Cora data}
\begin{figure*}[ht]
     \centering
     \includegraphics[width=0.49\textwidth]{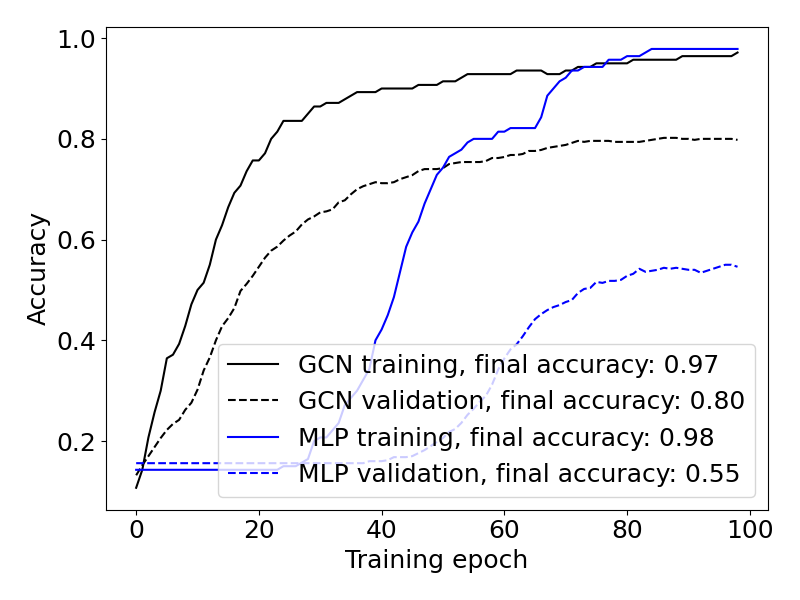}
     \includegraphics[width=0.48\textwidth]{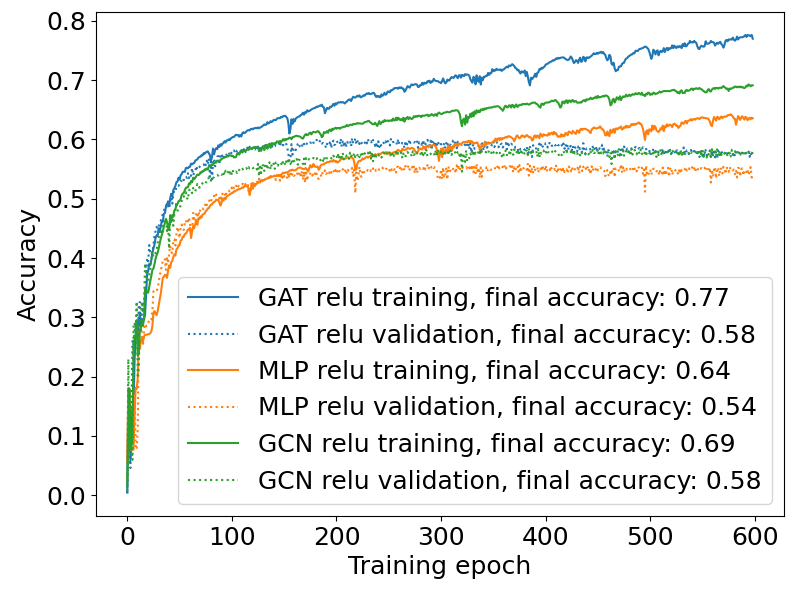}
%     \hspace*{-1.5cm} 
    \caption{The training (solid lines) and validation accuracies (dashed lines) during the fitting process. To the left is the Cora data with the MLP model in blue and GCN model in black. The to right is the arxiv data with the GAT model is blue, MLP model is orange and GCN model is green.}
    \label{fig:accuracy}
\end{figure*}

\figref{fig:accuracy} shows the training and validation accuracy during training. We see that both models obtain a training accuracy of almost 100\%, but the validation accuracy of the MLP model remains low, while the GCN structure enforces specific relationships in the model that appears to prevent some overfitting and consequently higher validation accuracy.

\begin{figure*}
%     \centering
     \includegraphics[width=0.49\textwidth]{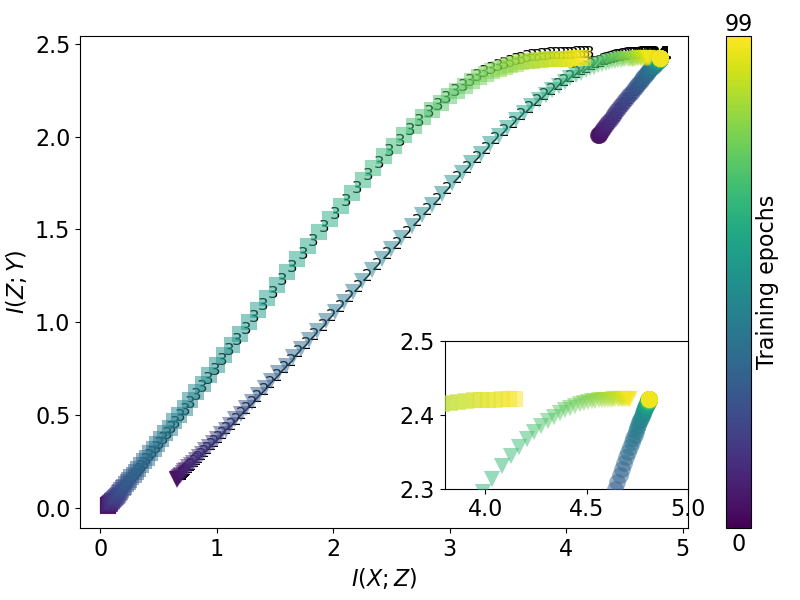}
     \includegraphics[width=0.49\textwidth]{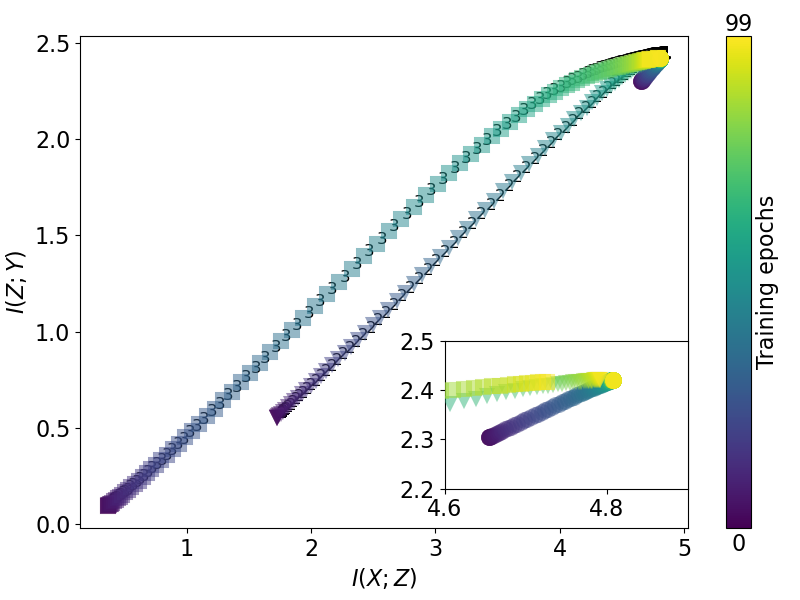} 
%     \hspace*{-1.5cm} 
    \caption{The information plane of the training process for MLP model (left) and GCN model (right) on the Cora citation dataset, with noise parameter $\sigma^2 =  0.1$ and \relu{} activations. The different symbols/numbers refer to layers in the networks ($\bigcirc$:1, $\triangle$:2, $\square$:3)
    whereas the colours refer to fitting epoch. The inserts provide a zoom-in of the convergence region. For the MLP the third layer (squares) ends up with smaller mutual information with X than the previous two layers (triangles and circles) adhering to the data processing inequality. Through its pooling operation, the GCN architecture is able to retain more information about $X$ than the MLP model, which is shown by the final layer (square) having noticeably higher mutual information about $X$ compared to its fully connected counterpart. }
    \label{fig:MI_plane_cora}
\end{figure*}

\figref{fig:MI_plane_cora} shows the information planes for the MLP and GCN models fitted to the Cora data. For the MLP model we see that the mutual information between input and the individual layers ($I(X, Z_i)$) and the individual layers and output ($I(Z_i, Y)$) increase with training, also referred to as the fitting phase. As expected, we do not observe any compression phase due to the choice of the asymmetric \relu{} activation function. For the MLP, the layers follow the data processing inequality (equation~\eqref{eqn:dpi-inequality}) with $I(X, Z_1) \geq I(X, Z_2) \geq I(X, Z_3)$ and $I(Z_1, Y) \geq I(Z_2,Y) \geq I( Z_3, Y)$. We note that even though the MLP model has less information about $X$ in its final layer compared to the GCN model, it is not able to use this in order to create generalized features which perform better on the validation set. Seen together with the gap between training and validation accuracy in \figref{fig:accuracy} this is a clear sign that the MLP architecture is unable to generalize well on this dataset.

For the GCN model we also observe that the mutual information between input and the individual layers ($I(X, Z_i)$) and the individual layers and output ($I(Z_i, Y)$) increase with training. By choosing $\sigma^2=0.1$ the data processing inequality is not violated, however for lower values of $\sigma^2$ the data processing inequality was violated for the GCN model. It is not always expected that the GCN architecture fulfills the data processing inequality. This is due to the successive representations $Z_i^1, Z_i^2$ drawing upon knowledge of the input data $X$ to generate the representations. Specifically for the GCN, they are obtained through an averaging of previous features from neighbouring nodes. This introduces an explicit dependence upon the structure of $X$ in the update step in equation~\eqref{eqn:graph-update-rule}, violating the data processing inequality. 

It is somewhat surprising that the GCN model retains more information about $X$ than the MLP model while at the same time performing better on the validation set. Usually one would seek to generalize away as much unnecessary variation in $X$ as possible in order to only retain variation related to predicting $Y$.  It seems the MLP architecture is bottlenecked by its ability to only convey information through matrix multiplications, whilst the GCN model alleviates this by also pooling each neighboring nodes' features in its propagation step.

\subsection{Results on arxiv data}
\figref{fig:accuracy} (right) shows the training and validation accuracy during training using \relu{} activations on the arxiv data. On the arxiv data, the models obtain training accuracies of 64-77\%, and validation accuracies of 54\% for the MLP model and 58\% for the two graph models\footnote{These are slightly worse than the official best fit models obtained on the data, but we have not tuned the individual models~\cite{2020arXiv200500687H} in order to keep them comparable}. The gap between training and validation accuracies indicate some overfitting, which is worst for the MLP. We assume this is due to the graph-based models enforcing specific relationships that helps with the generalization. The accuracies have very similar behaviour for the \tangh{} activations (not shown).

\begin{figure}
     \centering
     \includegraphics[width=0.45\textwidth]{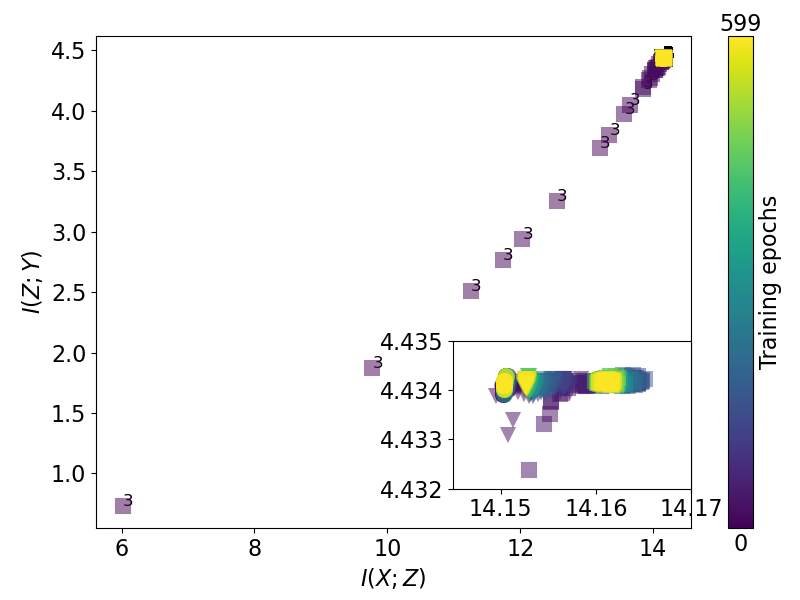}
     \includegraphics[width=0.45\textwidth]{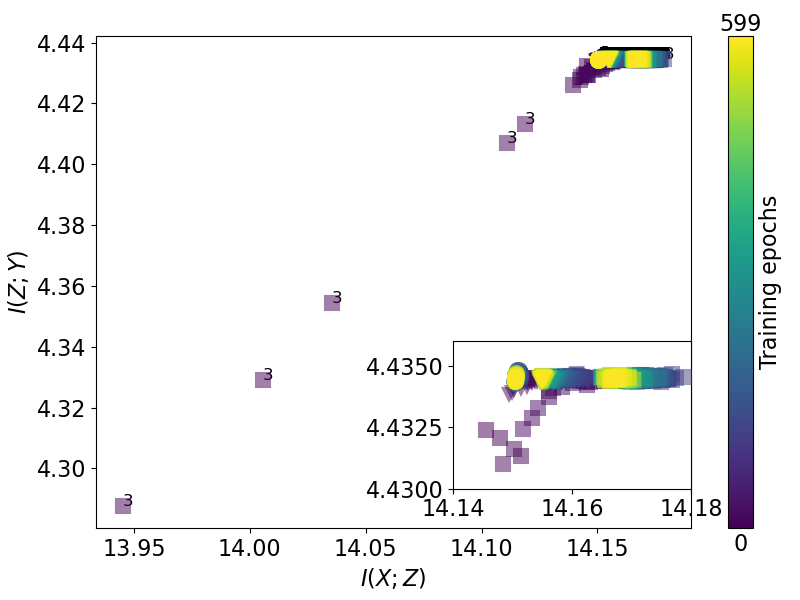} 
     \includegraphics[width=0.45\textwidth]{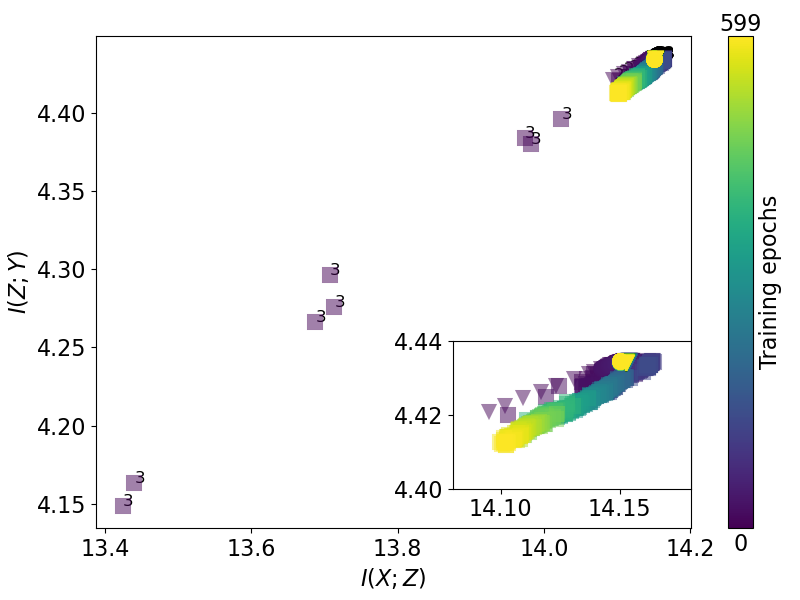} 
    \caption{The information plane of the training process for MLP model (top) and GCN model (middle) and GAT model (bottom) on the arxiv citation dataset, with noise parameter $\sigma^2 =  0.001$ and \relu{} activation functions. The different symbols/numbers refer to layers in the networks ($\bigcirc$:1, $\triangle$:2, $\square$:3) whereas the colours refer to fitting epoch. The inserts provide a zoom-in of the convergence region.
    }
    \label{fig:MI_plane_arxiv}
\end{figure}

\figref{fig:MI_plane_arxiv} shows the information planes for all three models fitted to the arxiv data with \relu{} activations. For all models we observe a rapid fitting phase with increasing mutual information along both axes. For the MLP model we see that all the layers ends up within a very small region and while the inset indicates that the ordering breaks the data processing inequality, this is most likely due to uncertainty in the estimation. We associate the closeness of the layers to a lack of compression during training and a poor generalization performance. The GCN and GAT show similar patterns, but with fewer iterations needed for the rapid increase of mutual information. The GAT performs a small compression of the last layer. In both cases, the closeness of the layers indicate low compression between the layers.

At the current stage, we are unable to draw general conclusions from the mutual information planes, but the presented framework will ease further study of the mutual information plane of inductively biased models.

\section{Conclusion}
We provide a framework for performing the information plane analysis by tracking the activations of a general PyTorch model. The work was partly motivated by the need for evaluating generalisation performance based on training data alone, and partly by the need for informed model architecture selection.
Being able to compare the mutual information plane across neural architectures is reliant on whether the architectures violate the data processing inequality. Ideally the data processing inequality would hold, but for more complicated architectures, and as shown in this work, that is not necessarily the case but the information plane can still provide insight in some cases. 
\typeout{} 
\bibliographystyle{abbrvnat}
\bibliography{references}

\end{document}